\def\year{2015}
\documentclass[letterpaper]{article}
\usepackage{aaai}
\usepackage{times}
\usepackage{helvet}
\usepackage{courier}
\usepackage{amsmath}

\usepackage{amsthm}
\usepackage{amssymb}
\usepackage{amsbsy}

\usepackage{url}
\usepackage{alltt}
\usepackage{subcaption}
\usepackage{comment}
\usepackage{epsfig}
\usepackage{algorithm}
\usepackage{algpseudocode}

\DeclareMathOperator*{\argmin}{arg\,min}

\theoremstyle{plain}

\theoremstyle{definition}

\newtheorem{example}{Example}

\def\st#1{\tt\small{#1}}
\def\textst#1{\text{\st#1}}

\title{
A Probabilistic Approach to Knowledge Translation
}

\author{
Shangpu Jiang \and Daniel Lowd \and Dejing Dou \\
Computer and Information Science \\
University of Oregon, USA\\
{\it\{shangpu,lowd,dou\}@cs.uoregon.edu}
}

\begin{document}

\maketitle

\begin{abstract}

In this paper, we focus on a novel knowledge reuse scenario where the
knowledge in the source schema needs to be translated to a semantically
heterogeneous target schema.  We refer to this task as ``knowledge
translation" (KT). Unlike data translation and transfer learning, KT
does not require any data from the source or target schema.  We adopt a
{\em probabilistic approach} to KT by representing the knowledge in the
source schema, the mapping between the source and target schemas, and
the resulting knowledge in the target schema all as probability
distributions, specially using  {\em Markov random fields} and {\em
Markov logic networks}.  Given the source knowledge and mappings, we use
standard learning and inference algorithms for probabilistic graphical
models to find an explicit probability distribution in the target schema
that minimizes the Kullback-Leibler divergence from the implicit
distribution.  This gives us a compact probabilistic model that
represents knowledge from the source schema as well as possible,
respecting the uncertainty in both the source knowledge and the mapping.
In experiments on both propositional and relational domains, we find
that the knowledge obtained by KT is comparable to other approaches that
require data, demonstrating that knowledge can be reused without data.

\end{abstract}

\section{Introduction}
\label{sec:intro}

Knowledge acquisition is a critical process for building predictive or
descriptive models for many applications. When domain expertise is
available, knowledge can be constructed manually. When enough
high-quality data is available, knowledge can be constructed
automatically using data mining or machine learning tools. Both
approaches can be difficult and expensive, so we would prefer to reuse
or transfer knowledge from one application or system to another whenever
possible. However, different applications or systems often have
different semantics, which makes knowledge reuse or transfer a
non-trivial task. 

As a motivating example, suppose a new credit card company without
historical data wants to use the classification model mined by a partner
credit card company to determine whether the applicants of the new
company are qualified or not. Since the two companies may use different
schemas to store their applicants' data (e.g., in one schema, we have
annual income recorded as a numerical attribute, while in the other, we
have salary as an attribute with discretized ranges), we cannot simply
reuse the old classifier. Due to privacy and scalability concerns, we
cannot transfer the collaborative company's data to the new schema
either. 
Therefore, we want to {\em translate} the classification model itself to
the new schema, {\em without using any data}.

In this paper, we propose {\em knowledge translation}  (KT) as a novel
solution to translate knowledge across conceptually similar but
semantically heterogeneous schemas or ontologies. For convenience, we
refer to them generically as ``schemas." 
As shown in the previous example, KT is useful in situations where data
translation\slash transfer is problematic due to privacy or scalability
concerns.

We formally define knowledge translation as the task of converting
knowledge $K_\mathcal{S}$ in source schema $\mathcal{S}$ to equivalent
knowledge $K_\mathcal{T}$ in target schema $\mathcal{T}$, where the
correspondence between the schemas is given by some mapping
$M_{\mathcal{S},\mathcal{T}}$. In general, one schema may have concepts
that are more general or specific than the other, so an exact
translation may not exist. We will therefore attempt to find the {\em
best} translation, acknowledging that the best translation may still be
a lossy approximation of the source knowledge.

We adopt a {\em probabilistic approach} to knowledge translation, in
which the knowledge in the source schema, the mapping between the source
and target schemas, and the resulting knowledge in the target schema are
all represented as probability distributions. This gives us a consistent
mathematical framework for handling uncertainty at every step in the
process. This uncertainty is clearly necessary when the source knowledge
is probabilistic, but it is also necessary when there is no exact
mapping between the schemas, or when the correct mapping is uncertain.
We propose to represent these probability distributions using {\em
Markov random fields}, for propositional (non-relational) domains, and
{\em Markov logic networks}, for relational domains.
Given probability distributions for both the source knowledge and the
schema mapping, we can combine them to define an implicit probability
distribution in the target schema. Our goal is to find an explicit
probability distribution in the target schema that is close to this
implicit distribution in terms of the Kullback-Leibler divergence. 

Our main contributions are:
\begin{itemize}

\item We formally define the problem of knowledge translation (KT),
which allows knowledge to be reused when data is unavailable.

\item We propose a novel {\em probabilistic} approach for KT by
combining probabilistic graphical models with schema mappings.

\item We implement an experimental KT system and evaluate it on two real
datasets 
We compare our data-free KT approach to baselines that use data from the
source or target schema and show that we can obtain comparable accuracy
without data.

\end{itemize}

The paper is organized as follows. We first summarize related work, such
as semantic integration, distributed data mining, and transfer learning,
and discuss their connections and distinctions with KT. We then show how
Markov random fields and Markov logic networks can represent knowledge
and mappings with uncertainty. Next, we present a variant of the
MRF\slash Markov logic learning algorithm to solve the problem of
knowledge translation. We then run experiments on synthetic and real
datasets. Finally, we make a conclusion.

\section{Related Work}
\label{sec:related}

In this section, we compare the task of knowledge translation with some
related work.

\paragraph{Semantic Integration}

Data integration and exchange (e.g., \cite{Lenzerini02}) are the mostly studied areas in semantic
integration. The main task of data integration and exchange is to answer
queries posed in terms of the global schema given source databases. The
standard semantics of global query answering is to return the tuples in
every possible database that is consistent with the global schema
constraints and the mapping, i.e., the set of {\em certain answers}.

A main difference between data integration\slash exchange and knowledge
translation (KT) is that KT has probabilistic semantics for the
translation process, that is, it defines a distribution of possible
worlds in the target schema, instead of focusing only on the tuples that
are in all the possible worlds (i.e., certain answers).

\paragraph{Distributed Data Mining}

Efforts in distributed data mining (DDM) (see surveys in
\cite{Park02,Caragea05}) have made considerable progress in mining
distributed data resources without putting data in a centralized
location. \cite{Caragea05} proposes a general DDM framework with 
two components: one sends statistical queries to local data sources, and
the other uses the returned statistics to revise the current partial 
hypothesis and generate further queries. 

Heterogeneous DDM \cite{Caragea05} also handles the semantic
heterogeneity between the global and local schemas, in particular, those
containing attributes with different granularities called Attribute
Value Taxonomy (AVT). Heterogeneous DDM requires local data resources
and their mappings to the global schema to translate the statistics of
queries. However, KT does not require data or statistics from either the
source or the target. Instead, KT uses mappings to translate the
generated/mined knowledge from the source directly. 

\paragraph{Transfer Learning}

Transfer learning (TL) has been a successful approach to knowledge
reuse. In traditional machine learning, only one domain and one task is
involved. When the amount of data is limited, it is desirable to use
data from related domains or tasks. As long as the source and target
data share some similarity (e.g., in the distribution or underlying
feature representation), such knowledge can be used as a ``prior'' for
the target task.

Most transfer learning work focuses on the homogeneous case in which the
source and target domain have identical attributes.  The main exceptions
are heterogeneous transfer learning \cite{Yang09} and relational
transfer learning (e.g., TAMAR \cite{Mihalkova07b}, deep
transfer \cite{Davis09}). Heterogeneous transfer learning deals with
different representations of the data (e.g., text and images of an
object). While it uses an implicit mapping of two feature spaces (e.g.,
through Flickr), KT uses an explicit mapping via FOL formulas.
Relational transfer learning deals with two analogous domains (e.g., in
movie and university domains, directors correspond to professors
). In contrast, KT focuses on a single domain with two different {\em
representations}. Moreover, relational transfer learning only handles
deterministic one-to-one matchings which can be inferred by using a
small amount of target data, while KT does not use any target data, and
relies on the provided explicit FOL mapping.

\paragraph{Deductive Knowledge Translation}

Deductive knowledge translation \cite{Dou11} essentially tries to solve
the same problem, but it only considers deterministic knowledge and
mappings. Our KT work can handle knowledge and mappings with
uncertainty, which is more general than the deterministic scenario
deductive knowledge translation \cite{Dou11} can handle. 

See Table~\ref{tab:related-work} for a summary of the similarities and
differences between our knowledge translation (KT) approach and
related work.

\begin{table}[!htb]
\centering
\footnotesize

\caption{Comparisons between KT and related work. We consider three aspects of a
task: whether data is available, what kind of knowledge patterns are supported,
and what kind of mapping is used.}

\label{tab:related-work}
\begin{tabular}{l | l l l}
\hline
& Data avail. & Knowledge & Mapping \\
\hline
Data integration & Source data & Query results & GLAV \\
Hetero. DDM & Source data & Propositional & AVT \\
Hetero. TL & Source/target & any & Implicit \\
Relational TL & Target data & SRL models & Matching \\
Deductive KT & No data & FOL & FOL \\
\hline
KT & No data & SRL models & SRL models \\
\hline
\end{tabular}

\end{table}

\newcommand{\bss}[0]{\boldsymbol}

\section{Probabilistic Representations of Knowledge and Mappings}

\label{sec:repr}

To translate knowledge from one schema to another, we must have a
representation of the knowledge and the mappings between the two
schemas. 
In many cases, knowledge and mappings are uncertain. For example, the
mined source knowledge could be a probabilistic model, such as a
Bayesian network. 
Mappings between two schemas may also be uncertain, either because a
perfect alignment of the concepts does not exist, or because there is
uncertainty about which alignment is the best. Therefore, we propose a
{\em probabilistic} approach to knowledge translation.

\subsection{Representation of Knowledge}


Our approach to knowledge translation requires that the source and
target knowledge are probability distributions represented as log-linear
models. In some cases, the source knowledge mined from the data may
already be represented as a log-linear model, such as a Bayesian network
used for fault diagnosis or Markov logic network modeling homophily in a
social network.  In other cases, we will need to convert the knowledge
into this representation.  

For mined knowledge represented as rules, including association rules, 
rule sets, and decision trees (which can be viewed as a special case 
of rule sets), we can construct a feature for each rule, with
a weight corresponding to the confidence or probability of the rule.
The rule weight has a closed-form solution based on the log odds that
the rule is correct:
\[
w_i = \log \frac{p(f_i)}{1-p(f_i)} - \log \frac{u(f_i)}{1-u(f_i)}
\]
where $p(f_i)$ is the probability or confidence of the $i$th rule or
formula and $u(f_i)$ is its probability under a uniform distribution.
Relational rules in an ontology can similarly be converted to a Markov
logic network by attaching weights representing their relative strengths
or confidences.

For linear classifiers, such as linear support vector machines or
perceptrons, we can substitute logistic regression, a probabilistic
linear classifier. 


In some cases, the knowledge we wish to translate takes the form of a
conditional probability distribution, $p(\bss{Y}|\bss{X})$, or a
predictive model that can be converted to a conditional probability
distribution. This includes decision trees, neural networks, and other
classifiers used in data mining and machine learning. The method we
propose will rely on a full joint probability distribution over all
variables. We can convert a conditional distribution into a joint
distribution by assuming some prior distribution over the evidence,
$p(\bss{X})$, such as a uniform distribution.

\subsection{Representation of Mappings}

The relationships between heterogeneous schemas can be represented as a
{\em mapping}. We use probabilistic models to represent mappings.
Consistent with the probabilistic representation of knowledge in a
database schema, the attributes are considered as random variables for
non-relational domains, and the attributes or relations are considered
as first-order random variables for relational domains. Let us denote
the variables in the source as $\boldsymbol{X} = \{X_1, ... , X_N\}$ and
those in the target as $\boldsymbol{X}' = \{X'_1, ..., X'_M\}$. A
mapping is the conditional distribution $p(\boldsymbol{X}'|\boldsymbol{X})$.

In real cases, a mapping is often represented as a set of
source-to-target correspondences
\[
\{p(\boldsymbol{C}'_i|\boldsymbol{C}_i), i=1,...,I\}
\]
where $\boldsymbol{C}_i \subset \boldsymbol{X}$ and $\boldsymbol{C}'_i \subset
\boldsymbol{X}'$ are sets of variables in the source and target respectively.
For the credit card company example, a mapping between the two schemas may include
the correspondences of ``age'' and ``age," ``salary'' and ``annual income," etc.

In order to obtain a global mapping between the source and target
schemas using the local correspondences, we make the following two
assumptions:

\begin{enumerate}

\item $p(\boldsymbol{C}'_i \cup \boldsymbol{C}'_j | \boldsymbol{X}) =
p(\boldsymbol{C}'_i | \boldsymbol{X}) p(\boldsymbol{C}'_j |
\boldsymbol{X})$, or, $\boldsymbol{C}'_i\perp \boldsymbol{C}'_j |
\boldsymbol{X}$;

\item $p(\boldsymbol{C}'_i | \boldsymbol{X}) =
p(\boldsymbol{C}'_i|\boldsymbol{C}_i)$.

\end{enumerate}
From these two assumptions, it follows that:
\[
p(\boldsymbol{X}'|\boldsymbol{X}) 
= \prod_i p(\boldsymbol{C}'_i|\boldsymbol{X}) 
= \prod_i p(\boldsymbol{C}'_i|\boldsymbol{C}_i)
\]

Note that these assumptions are not always correct, but they provide a
good approximation of the global mapping when it is not available.


The weight of each formula can be estimated with the log-odds. For example, we
define a probabilistic source-to-target correspondence as $q_\mathcal{S}
\rightarrow_p q_\mathcal{T}$, where $q_\mathcal{S}$ and $q_\mathcal{T}$ are
queries (i.e., logical formulas) of source and target schemas or ontologies,
and $\rightarrow_p$ has probabilistic semantics:
\[
\text{Pr}(q_\mathcal{T} | q_\mathcal{S}) = p
\]

\begin{example}[Class correspondence]

If $x$ is a graduate student, then $x$ is a student and older than 24 with
probability 0.9, and vice versa.
\begin{align*}
\textst{Grad}(x) \rightarrow_{0.9} \textst{Student}(x) \land \textst{Age}(x,y) \land (y \ge 24) \\
\textst{Grad}(x) \leftarrow_{0.9} \textst{Student}(x) \land \textst{Age}(x,y) \land (y \ge 24)
\end{align*}
This can be converted to
\begin{align*}
2.2 \qquad \textst{Grad}(x) \rightarrow (\textst{Student}(x) \land \textst{Age}(x, y) \land (y \ge 24)) \\
2.2 \qquad \textst{Grad}(x) \leftarrow (\textst{Student}(x) \land \textst{Age}(x, y) \land (y \ge 24))
\end{align*}

\end{example}

%


\section{Knowledge Translation}
\label{sec:method}

In this section, we formalize the task of knowledge translation (KT) and
propose a solution to this task. We have the source knowledge
represented as a probabilistic model $p(\boldsymbol{X}) = p(X_1, ...
X_n)$ and a probabilistic mapping $P(\boldsymbol{X}'|\boldsymbol{X})$.
The probabilistic model in the target schema can be computed as
\begin{align}
p(\boldsymbol{X}') = \sum_{\boldsymbol{X}} p(\boldsymbol{X}) p(\boldsymbol{X}' | \boldsymbol{X}) 
= \sum_{\boldsymbol{X}} p(\boldsymbol{X}) \prod_i p(\boldsymbol{C}'_i | \boldsymbol{C}_i) \label{eq:knowl_trans}
\end{align}
Our goal is to find a {\em compact} probabilistic model in the target schema
(i.e., the target knowledge) without using any source variables as latent
variables. This requirement is due to both efficiency (when the knowledge is
being used) and understandability consideration.

We also use a log-linear model $q(\boldsymbol{X}')$ to represent this
compact model. A straight-forward objective is to minimize the
Kullback-Leibler divergence 
\begin{align}
q^* & = \argmin_{q} D_\text{KL} \left[p(\boldsymbol{X}') \| q(\boldsymbol{X}')\right] \nonumber \\
& = \argmin_{q} - \sum_{\boldsymbol{X}'} p(\boldsymbol{X}') \log q(\boldsymbol{X}') \label{eq:kl-div}
\end{align}

The joint distribution $p(\boldsymbol{X}, \boldsymbol{X}')$ is also a
log-linear model (see Equation~\ref{eq:knowl_trans}). The weights for
a local correspondence can be computed as:
\begin{align*}
\bar{\theta}(\boldsymbol{C}_i, \boldsymbol{C}'_i) = \log p(\boldsymbol{C}'_i | \boldsymbol{C}_i) 
= \log \frac{\exp (\theta (\boldsymbol{C}_i, \boldsymbol{C}'_i))}{\sum_{\boldsymbol{C}'_i} \exp {\theta (\boldsymbol{C}_i, \boldsymbol{C}'_i)}}
\end{align*}
where $\theta (\boldsymbol{C}_i, \boldsymbol{C}'_i)$ are the weights of the
correspondence in the probabilistic mapping model. The computation of
$p(\boldsymbol{X}')$ is therefore a standard inference task of the joint model
$p(\boldsymbol{X}, \boldsymbol{X}')$.

\paragraph{Parameter Learning}

The parameters of the target log-linear model that minimizes
Equation~\ref{eq:kl-div} can be computed via standard optimization
algorithms. A simple way to compute the objective is sampling: we first
generate a sample from the source $p(\boldsymbol{X})$, and then generate
a sample of $\boldsymbol{X}'$ from $p(\boldsymbol{X}'|\boldsymbol{X})$
conditioned on the source sample. In the relational domain (with Markov
logic or other statistical relational models), each sample instance is a
database, and we need to first decide the number of constants and create
a set of ground variables with these constants.

\paragraph{Structure Learning}

The structure of the target knowledge can also be learned via standard
structure learning algorithms for Markov random fields or Markov logic
networks. An alternative approach is to use heuristics to generate the
structure first. For deterministic one-to-one correspondences, the
independences in the target schema are the same as those in the source
schema up to renaming. If the correspondences are non-deterministic, we
may have less independences in the target schema, and we could
have an extremely complex model with large cliques. Nonetheless, in
realistic scenarios, the correspondences in a mapping are usually
deterministic or nearly deterministic. Therefore, it is reasonable to
pretend they are deterministic while inferring the target structure. In
this way we trade off between the complexity and accuracy of the target
knowledge. 

First of all, for Markov logic, we use first-order cliques instead of
formulas as the source structure, so that it is consistent with the
propositional case. We show the pseudocode of the structure translation
in Algorithm~\ref{alg:struct-learn}. It is considered as a structure
learning process. The first step (Line 1-8) is to remove the variables
that do not have a correspondence in the target schema. This can be done
by standard variable elimination \cite{Koller09,Poole03} without
calculating parameters. However, exact variable elimination may create
very large cliques and be very expensive, especially in Markov logic in
the relational domains.  Therefore, we approximate it by only merging
two cliques at a time. For relational case, the merging involves a
first-order unification operation \cite{Russell03,Poole03}. When
multiple most general unifiers exist, we simply include all the
resulting new cliques. In the second step (Line 9-16), we replace each
variable with the corresponding variables in the target schema. This
also involves first-order unification in the relational case. If there
are many-to-many correspondences, we may generate multiple target
cliques from one source clique.

\begin{algorithm}

\caption{Structure Translation (MRFs or MLNs)}
\label{alg:struct-learn}

\textbf{Input:} The source schema $\mathcal{S}$, source structure
(propositional or first-order cliques) $\Phi = \{ \phi_i \}$, and
mapping $\mathcal{M}$.

\textbf{Output:} The target structure $\Phi'_{\mathcal{M}}$.

\begin{algorithmic}[1]
\For{\textbf{each} variable (or first-order predicate) $P \in \mathcal{S}$ that does not appear in $\mathcal{M}$}
    \State Let $\Phi_P$ denote all the cliques containing $P$
    \State Remove $\Phi_P$ from $\Phi$
    \For{\textbf{each} pair of cliques in $\Phi_P$}
    	\State Merge the two cliques and remove $P$
		\State Insert the resulting clique back to $\Phi$
    \EndFor
\EndFor

\For{\textbf{each} clique $\phi \in \Phi$}
	\For{\textbf{each} variable $P$ in $\phi$}
		\State Let $P'_{\mathcal{M}}$ be all possible correspondences of $P$
	\EndFor
	\State Let $\phi'_{\mathcal{M}}$ denote all possible correspondences of $\phi$
    \State $\phi'_{\mathcal{M}} \leftarrow$ Cartesian product of $P'_{\mathcal{M}}$
    \State Add $\phi'_{\mathcal{M}}$ to $\Phi'_{\mathcal{M}}$
\EndFor
\end{algorithmic}

\end{algorithm}

\begin{example}

Given the source Markov logic:
\begin{align*}
\textst{Grad}(x) & \rightarrow \textst{AgeOver25}(x) \\
\textst{AgeOver25}(x) & \rightarrow \textst{GoodCredit}(x)
\end{align*}
and the mapping:
\begin{align*}
2.2 \qquad & \textst{Grad}(x) \lor \textst{Undergrad}(x) \leftrightarrow \textst{Student}(x) \\
3.0 \qquad & \textst{GoodCredit}(x) \leftrightarrow \textst{HighCreditScore}(x)
\end{align*}

We first eliminate $\textst{AgeOver25}(x)$ from the source structure
because it does not occur in the mapping, and we get a new clique
\[
\{\textst{Grad}(x), \textst{GoodCredit}(x)\}
\]
Then we translate the clique based on the mapping, which gives
\[
\{\textst{Student}(x), \textst{HighCreditScore}(x)\}
\]

\end{example}

%

\section{Experiments}
\label{sec:exp}

To evaluate our methods, we created two knowledge translation tasks: one
on a non-relational domain (NBA) and one on a relational domain
(University). In each knowledge translation task, we have 2 different
database schemas as the source and target schemas and a dataset for each
schema. The input of a knowledge translation system is the source
knowledge and the mapping between the source and target schema. We
obtained the source knowledge (i.e., a probabilistic model in the
source) by performing a common learning algorithm on the source dataset,
and created the probabilistic schema mapping manually. The output of a
knowledge translation system is the target knowledge (i.e., a
probabilistic model in terms of the target schema). 

\subsection{Methods and Baselines}


We evaluate three different versions of our proposed probabilistic
knowledge translation approach described in the previous section.  All
of them use the source knowledge and probabilistic mapping to generate a
sampled approximation of the distribution in the target schema, and all
of them use these samples to learn an explicit distribution in the
target schema. The difference between them is their approach to
knowledge structure. \textbf{LS-$K_S$} (``learned structure") learns the
structure directly from the samples, which is the most flexible
approach.
\textbf{TS-$K_S$} (``translated structure") uses a heuristic translation
of the structure from the source knowledge base.
\textbf{ES-$K_S$} (``empty structure") is a simple baseline in which the
target knowledge base is limited to a marginal distribution.

We also compare to several baselines that make use of additional data.
When there is data $D_S$ in the source schema, we can use the
probabilistic mapping to translate it to the target schema and learn
models from the translated source data. \textbf{LS-$D_S$} and
\textbf{MS-$D_S$} learn models from translated source data, using
learned and manually specified structures, respectively.
When there is data $D_T$ in the target schema, we can learn from this
data directly. \textbf{LS-$D_T$} and \textbf{MS-$D_T$} learn models from
target data with learned and manually specified structures respectively.
These methods represent an unrealistic ``best case" since they use data
that is typically unavailable in knowledge translation tasks. 



We evaluate our knowledge translation methods according to two criteria: the
pseudo-log-likelihood (PLL) on the held-out {\em target data}, and PLL
on the held-out {\em translated source data}. The advantage of the
second measure is that it controls for differences between the source
and target distributions.
For relational domains, we use weighted pseudo-log-likelihood (WPLL),
where for each predicate $r$, the PLL of each of its groundings is
weighted by the $c_r = 1 / g_r$, where $g_r$ is the number of its
groundings. 

\subsection{Non-Relational Domain (NBA)}

We collected information on basketball players in the National
Basketball Association (NBA) from two websites, the NBA official
website \texttt{nba} (as the source schema) and the Yahoo NBA website
\texttt{yahoo} (as the target schema). The schemas of these two
datasets both have the name, height, weight, position and team of each
player. In these schemas, the values of position have a different granularity.
Also, in \texttt{nba}, we discretize
height and weight into 5 equal-width ranges.  In \texttt{yahoo}, we
discretize them into 5 equal-frequency ranges (in order to make the mapping more challenging).
The
correspondences of these attributes are originally unit conversion formulas, e.g.,
\[
h' = h \times 39.3701
\]
After we discretize these attributes, we calculate the correspondence
distribution of the ranges by making a simple assumption that
each value range is uniformly distributed, e.g.,
\begin{align*}
p(h' \in (73.5,76.5] | h \in (1.858,1.966]) & = 0.706
\end{align*}

We used the Libra Toolkit\footnote{\url{http://libra.cs.uoregon.edu/}}
for creating the source knowledge and for performing the learning and
inference subroutines required by the different knowledge translation
approaches.
We first left out 1/5 of the data instances in the source and target
dataset as the testing sets. For the remaining source dataset, we used
the decision tree structure learning (DTSL) \cite{lowd14} to learn the
source knowledge. We used standard 4-fold cross validation to determine
the parameters of the learning algorithm. The parameters include
$\kappa$, prior, and mincount for decision tree learning, and $l_2$ for
weight learning.

We use Gibbs sampling for the sampling algorithm in the knowledge
translation approaches. For LS-$K_S$ and TS-$K_S$, we draw $N$ samples
from the source knowledge probability distribution. 
We then use the probabilistic mapping to draw 1 target sample for each
source sample. For LS-$D_S$, suppose we have $N_S$ instances in the
source dataset. We use the probabilistic mapping to draw $N / N_S$
target samples for each source instance, such that the total number of
target instances is also $N$.  

LS-$K_S$ and TS-$K_S$ both perform weight learning with an $l_2$
prior. For structure translation with TS-$K_S$, we only translate
features for which the absolute value of the weight is greater than a
threshold $\theta$.  These two parameters are tuned with cross
validation over a partition of the samples. 

\begin{figure*}
\centering
\hspace{-0.3in} \includegraphics[scale=0.3]{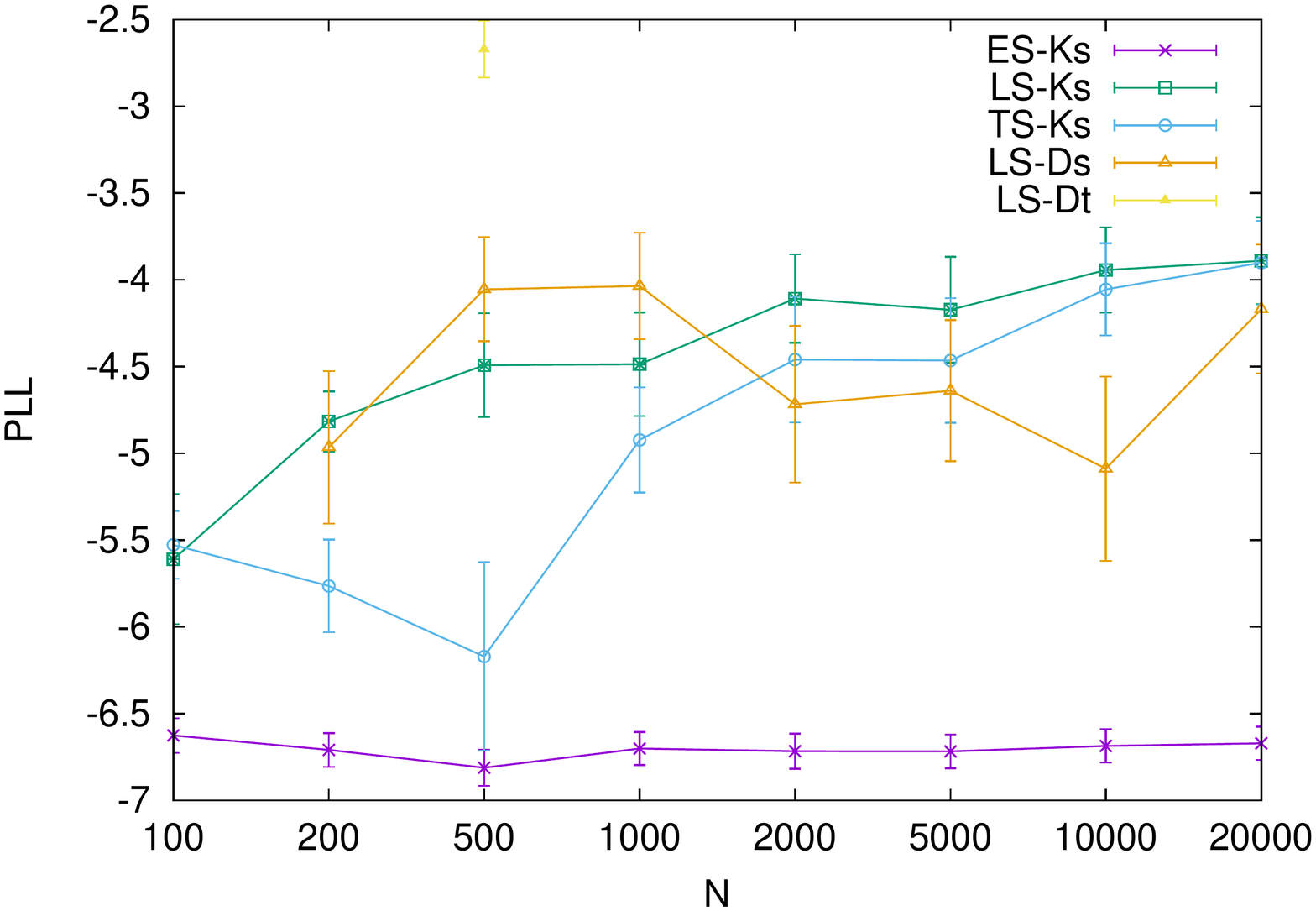}
\hspace{-0.3in} \includegraphics[scale=0.3]{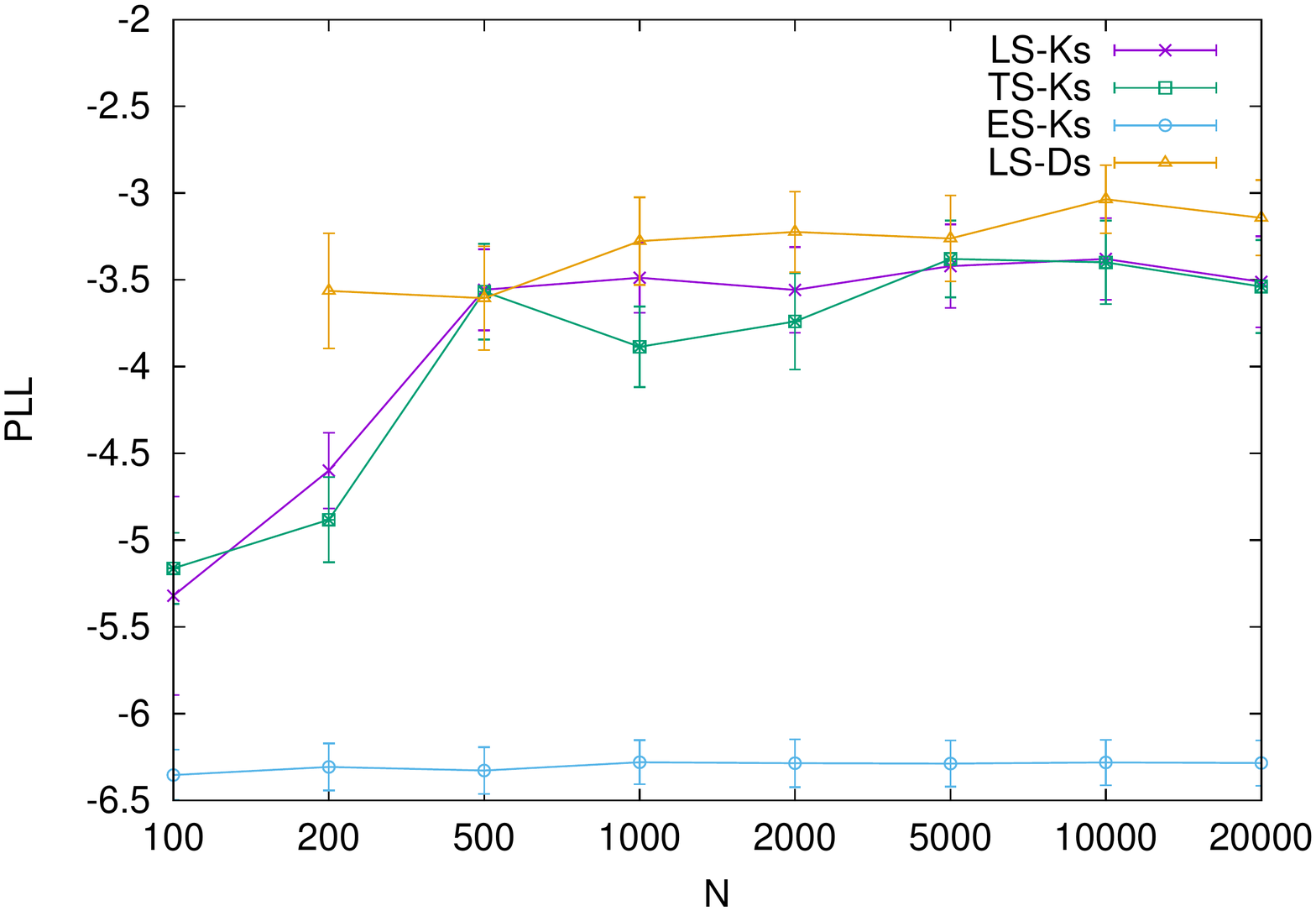}
\caption{PLL for KT methods and baselines on target data (left) and translated source data (right) in the NBA domain.}
\label{fig:lc1}
\end{figure*}

See Figures~\ref{fig:lc1} for learning curves comparing our methods to
the baselines. 
We see that translated knowledge (LS-$K_S$ and TS-$K_S$) is as accurate
as knowledge learned from translated source data (LS-$D_S$) on both the
target data and the translated source data. This confirms that {\em KT
can be as accurate as data translation}, but with the advantage of not
requiring any data.  We do not see a large difference between learning
the structure (LS-$K_S$) and heuristically translating the structure
(TS-$K_S$).  As expected, the model learned directly on the target data
(LS-$D_T$) has the best PLL on the target data, since it could observe
the target distribution directly.



\subsection{Relational Domain (University)}

We use the \texttt{UW-CSE}
dataset\footnote{\url{http://alchemy.cs.washington.edu/data/uw-cse/}.}
and the \texttt{UO-CIS} dataset which we collected from the Computer and
Information Science Department of the University of Oregon. The
\texttt{UW-CSE} dataset was introduced by Richardson and
Domingos \cite{Richardson06} and is widely used in statistical
relational learning research. In this University domain, we have
concepts such as persons, courses, and publications; attributes such as
PhD student stage and course level; and relations such as advise, teach,
and author.  The schemas of the two databases differ in their
granularities of concepts and attribute values. For example, 
\texttt{UW-CSE} graduate courses are marked as level 500, while
\texttt{UO-CIS} has both graduate courses at level 600 and combined
undegraduate/graduate courses at level 4/500.

Our methods in this relational domain are similar to those in the
non-relational domain.  We use
Alchemy~\footnote{\url{http://alchemy.cs.washington.edu/alchemy1.html}}
for learning and inference in Markov logic networks. We obtain the
source knowledge by manually creating the formulas in the source
schema and then using the source data to learn the weights.

We use MC-SAT \cite{Poon06} as the sampling algorithm for these
experiments. 
Since the behavior of a Markov logic network is highly sensitive to the
number of constants, we want to keep the number of constants similar to
the original dataset from which the model is learned. We set the number
of constants of each type to be the average number over all training
databases, multiplied by a scalar $\frac{1}{2}$ for more efficient
inference.
For methods based on $K_S$, we draw $N$ samples from the source
distribution and 1 target sample from each source sample and the
mapping. For methods based on $D_S$, we draw $N$ samples based on the
mapping. Here $N$ does not have to be large, because each sample
instance of a relational domain is itself a database.  We set $N$ to 1,
2 and 5 in our experiments.
We set the $l_2$ prior for weight learning to 10, based on
cross-validation over samples. 

The results are shown in Table~\ref{tab:univ1}.  In general, learning
MLN structure (LS-$K_S$ and LS-$D_S$) did not work as well as their
counterparts with manually specified structures (MS-$K_S$ and MS-$D_S$).
From a single sample, the translated source data and manually specified
structure (MS-$D_S$) were more effective than knowledge translation with
translated structure (TS-$K_S$). However, as we increase the number of
samples, the performance of TS-$K_S$ improves substantially.  With 5
samples, the performance of TS-$K_S$ becomes competitive with that of
MS-$D_S$, again demonstrating that knowledge translation can achieve
comparable results to data translation but without data. When evaluated
on translated source data, TS-$K_S$ shows the same trend of improving
with the number of samples, but its performance with 5 relational
samples is slightly worse than MS-$D_S$.

\begin{table}

\small
\centering

\caption{Evaluation on the target dataset (left) and translated source
dataset (right) for the university domain. N/A means it takes too much
time to run.}

\label{tab:univ1}

\begin{tabular}{c | c c c || c c c}
\hline
Method & \multicolumn{3}{c||}{WPLL on target} & \multicolumn{3}{c}{WPLL on source} \\
\hline
\# Samples & 1 & 2 & 5 & 1 & 2 & 5\\
\hline
\textbf{ES-$K_S$} & -3.77 & -3.76 & -3.83  & -3.54 & -3.44 & -3.39 \\
\textbf{LS-$K_S$} & -12.07 & -3.82 & -3.48  & -9.19 & -3.72 & -1.51 \\
\textbf{TS-$K_S$} & -2.51 & -2.80 & -1.79 & -2.05 & -2.10 & -0.97 \\
\textbf{LS-$D_S$} & -3.70 & -3.01 & N/A  & -1.23 & -1.23 & N/A \\
\textbf{MS-$D_S$} & -1.94 & -1.91 & -1.76  & -1.22 & -0.93 & -0.61 \\
\textbf{LS-$D_T$} & -1.33 & & & & & \\
\textbf{MS-$D_T$} & -1.18 & & & & & \\
\hline
\end{tabular}

\end{table}

\section{Conclusion}
\label{sec:concl}

Knowledge translation is an important task towards knowledge reuse where
the knowledge in the source schema needs to be translated to a
semantically heterogeneous target schema. Different from data
integration and transfer learning, knowledge translation focuses on the
scenario that the data may not be available in both the source and
target.  We propose a novel probabilistic approach for knowledge
translation by combining probabilistic graphical models with schema
mappings. We have implemented an experimental knowledge translation
system and evaluated it on two real datasets for different prediction
tasks. The results and comparison with baselines show that our approach
can obtain comparable accuracy without data.

The proposed log-linear models, such as Markov random fields and Markov
logic networks, already cover most of common types of knowledge used in
data mining. In the future work, we will extend our approach to the
knowledge types which are harder to represent as log-linear models, such
as SVMs and nearest neighbor classifiers. It might require a specialized
probabilistic representation. 


\paragraph{Acknowledgement} This research is funded by NSF grant IIS-1118050.

%
\bibliographystyle{aaai}

\bibliography{skti}  
%
%
\end{document}